\def\year{2020}\relax
\documentclass[letterpaper]{article} 
\usepackage{aaai20}  
\usepackage{times} 
\usepackage{helvet} 
\usepackage{courier}  
\usepackage[hyphens]{url}  
\usepackage{graphicx} 
\usepackage{threeparttable}
\usepackage{booktabs}
\usepackage{graphicx}  
\title{Towards Preference Learning for Autonomous Ground Robot Navigation Tasks}
\author{Cory Hayes, Matthew Marge\\ 
CCDC Army Research Laboratory\\
2800 Powder Mill Rd.\\
Adelphi, MD 20783\\
\{cory.j.hayes4.civ, matthew.r.marge.civ\}@mail.mil 
}
 
\begin{document}

\maketitle
\begin{abstract}
We are interested in the design of autonomous robot behaviors that learn the preferences of users over continued interactions, with the goal of efficiently executing navigation behaviors in a way that the user expects.
In this paper, we discuss our work in progress to modify a general model for robot navigation behaviors in an exploration task on a per-user basis using preference-based reinforcement learning.
The novel contribution of this approach is that it combines reinforcement learning, motion planning, and natural language processing to allow an autonomous agent to learn from sustained dialogue with a human teammate as opposed to one-off instructions.
\end{abstract}

\section{Introduction}
For robots to effectively team with humans, they will need to be considered trustworthy.
Two critical elements of robot trustworthiness are reliability and predictability \cite{Desai12}.
In this paper, we describe our work in progress investigating learning approaches for autonomous robot teammates that improve their reliability and predictability by executing navigation behaviors that are tailored to what their human teammates expect them to do.
The context for this work is improving a robot's navigation capabilities over the course of an interactive dialogue with a human teammate.
The goal of this research is to determine whether incorporating user preferences for navigation leads to more efficient interactions over time.
We hypothesize that this is possible, and improvements could be measured by (1) the overall amount of time required to complete a collaborative exploration task and (2) a reduction in the number of intervening clarification commands given to a robot during back-and-forth dialogue with a human teammate.

For cognitively intensive tasks that require a team, one or more human operators may be faced with a great deal of responsibility, needing to make quick decisions as the environment changes constantly.
Robots that must be instructed at each step only increase user burden, but on the opposite end of the spectrum, human teammates may not be able to predict the actions of robots that fully act independently and this in turn negatively impacts trust \cite{hancock,desaikania}.
A robot teammate that can learn user preferences could allow it to predict user actions by leveraging prior interaction history.
This would also allow it to play a more proactive role in accomplishing a team task instead of being purely reactive, and lead to more efficient missions.
The need for active robot team members that can appropriately shift between directly following orders and making effective decisions on their own motivates the design of robots that can infer human intent.

In order to acquire information about an environment in the quickest and most efficient manner, an intelligent robot teammate must be able to operate in a way that meets expectations.
This requirement must be met both when the robot is within the proximity of its human teammates as well as when it is remote, needing to be able to relay relevant information in the latter case.
For example, if a robot is instructed to survey an interior area and send images, there are many possible ways for this task to be accomplished.
At the same time, there may be variations in how the human operator wants the task to be carried out.
One user may want to make decisions as quickly as possible by having the robot move to one position that shows as much of a room as possible, take an image, and then move to a new position to await further instructions.
Another user may be more meticulous and want the robot to take multiple pictures from different perspectives in one area to carefully decide on the next course of action.
A robot that does not have the capability to pick up on user preferences by leveraging prior interaction history would require the user to constantly step in and make minor adjustments to its behavior if its default navigation style was not suitable.

\begin{figure}
\includegraphics[scale=0.38]{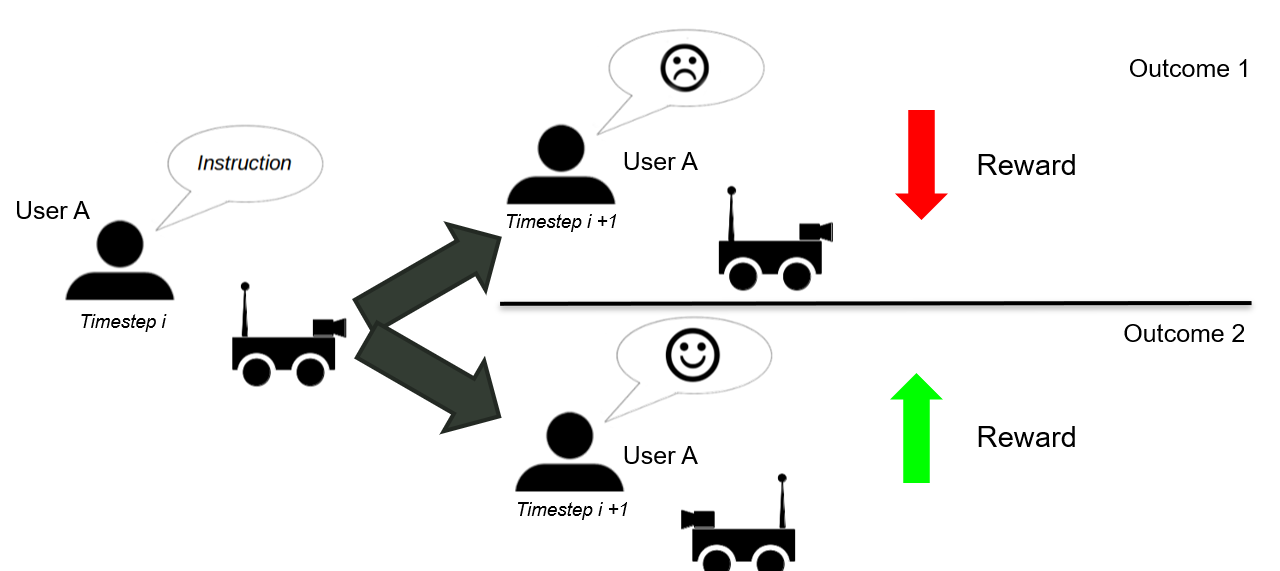}
\caption{The planned capability of our preference-learning navigation algorithm. User A gives a verbal instruction to a robot who then executes the behavior, and in return the user either has a negative (top right) or positive (bottom right) response to the robot's actions to be used as reward signals.}
\label{fig:sentiment}
\end{figure}

The overall goal of this work is to add a motion planning component for autonomous robot navigation that takes into account the user's preference for how navigation should be carried out.
We seek to combine human-robot interaction and preference-based reinforcement learning so that an autonomous agent can learn users' preferences based on how the user has reacted to the robot executing previous instructions; see Figure~\ref{fig:sentiment} for a simplified illustration of the intended interaction outcome.
We hypothesize that navigation models that adapt to user preferences will lead to more efficient interactions with less user burden.

\section{Background and Related Work}
\subsection{Human-Robot Dialogue Experiments}
This research is motivated by a series of experiments \cite{bonialAAAI,margeNLP,lukinACL} we conducted that collected robot-directed dialogue in a collaborative exploration task with humans.
In the current work, we build upon its collected data and methodology to capture user preferences.
In the experiments, participants would direct a remotely-located robot through unfamiliar indoor environments to gather information about the surroundings.
Participants could derive context strictly from observing sensory data sent by the robot: streaming laser scans of its map, images the robot could send upon request, and language responses in the form of status updates and clarification requests.
Across most of the experiments we have used a Wizard-of-Oz approach where one researcher substitutes for the automated dialogue manager that processes user speech, and a second researcher  substitutes for the autonomous navigation component upon receiving the processed request from the dialogue manager~\cite{Marge2016}.
We are currently in the stage where the dialogue manager is now fully automated; the proposed user preference learning capability is its own line of research that may eventually be integrated into the actual autonomous navigation component in the future.

\begin{table*}
\small
  \caption{Navigation Instruction Types}
  \centering
  \begin{threeparttable}
  \begin{tabular}{ccl}
    \toprule
    Instruction Type&Example Instruction\\
    \midrule
    Approach an entrance & ``Go to the doorway''\\
    Enter an unexplored room & ``Go into the room in front of you''\\
    Approach an object's personal space & ``Move forward until you reach the crate''\\
    Navigate unexplored areas & ``Go to the end of the hallway''\\
    Handle object obstruction & ``Move around the cone''\\
    Follow direction-only instruction & ``Go to the left''\\
    Follow direction \& orientation instruction & ``Head three feet north''\\
    Fulfill an impractical task & ``Move forward eight feet''\tnote{1}\\
  \bottomrule
\end{tabular}
\begin{tablenotes}
 \item{1} This instruction is impractical due to the robot's proximity to a wall.
\end{tablenotes}
\end{threeparttable}
\label{tab:instructions}
\end{table*}

A key observation from these previous experiments is that natural language commands can elicit many possible navigation paths for a robot, based on immediate physical context, dialogue history, and individual user preferences.
We have made a previous effort into understanding participant preferences that steer away from our preconceived notions of navigation behaviors \cite{moolchandani}, and uncovered eight types of navigation instructions as shown in Table~\ref{tab:instructions}; the proposed work here is an extension to actively learn these preferences in a setup similar to our existing studies.

\subsection{Preference-Based Reinforcement Learning}
A robot that can incorporate feedback from both the environment and participants as it takes actions should be able to learn to act intelligently and in a way that best suits the user.
Reinforcement learning enables an agent to learn how to achieve desired behavior by having it carry out actions and receive feedback from the environment that informs it on how valuable the action was towards achieving this goal.
It is an ideal technique to use for problems where one cannot explicitly provide the specific actions and sequences needed to accomplish a goal for all possible situations because it can evaluate if the actions taken bring the agent closer to the desired behavior (e.g., driving a car, walking, etc.).
This applies to our problem of capturing user preferences for robot navigation because it is not possible for participants to list all of the specific actions that a robot should take for any given environment in advance, especially for unfamiliar ones.
Our research problem is similar to the one addressed by the TAMER work \cite{knox12} where human input is used to speed up the guidance of agent behavior, but we focus on using verbal input on guiding robot navigation behavior.

One limitation of reinforcement learning is the reliance on arbitrarily setting numerical rewards that the agent seeks to maximize.
This dependency can result in unexpected agent behavior \cite{Wirth17} where the agent may maximize the reward but not actually carry out the task.
It can also make determining the correct policy more complicated by having an imbalance between defining the goal and guiding the agent to the correct solution, avoiding infinite rewards, and creating difficulties in determining the correct policy for tasks with multiple reward signals.
Another concern is that if the reward signal is affected at all by a non-expert user who may not know the optimal behavior desired, the learning process may not be able to occur successfully.
This limitation has been addressed in other works (e.g., \citeauthor{ziebart} \citeyear{ziebart}) that relax the requirement for the user to always behave optimally, but the limitation of the learning agent not being able to perform better than the user remains.

Preference-based reinforcement learning is a subfield of reinforcement learning where the reward signals use relative utility values to counteract the aforementioned problems when using absolute values.
It is centered on pairwise comparisons that express preferences between different data objects or labels for the objects.
Preference-based reinforcement learning is represented as a Markov Decision Process With Preference (MDPP) as a sextuple (S,A,$\mu$,$\delta$,$\gamma$,$\rho$), with state space S, action space A(s), initial state set $\mu$(s), state transition model $\delta$(s'$\vert$s,a), discount factor $\gamma$ $\in$ [0,1), and finally $\rho$($\tau$$_1$ $\succ$ $\tau$$_2$) to show the probability of preferring a trajectory $\tau$ = \{s$_0$,a$_0$,s$_1$,a$_1$,...,s$_{n-1}$,a$_{n-1}$,s$_n$\} over another.
Since the trajectories are not provided by the users and they do not require fixed reward values, the learning algorithm is able to provide output trajectories that can be better than what the user can demonstrate themselves, which is an applicable advantage for our robot navigation task scenario given the general population's overall unfamiliarity with robots \cite{riek2012}.

\section{Approach}
The overall goal of this work is to develop a machine learning component that can learn user preferences for navigation through feedback.
For this task, the feedback is user speech that is converted to formatted text that can be processed by our navigation algorithm.
We use the Robot Operating System (ROS) and the Gazebo simulator as our experimental testing environment.
For the initial stage of this experiment, we have opted to go with simulation for initial proof-of-concept and more rapid development compared to the same process on a physical platform; however the ultimate objective for this research is to implement our preference learning navigation algorithm on physical ground robot platforms.

\subsection{Experimental Setup}
Similar to our previous human-robot dialogue experiments, participants will be given an exploration task to collect information about an unfamiliar environment via LIDAR and RGB camera sensors on a virtual robot, which will require them to verbally instruct the robot.
The user will be tasked with gathering information about an environment to answer questions within a certain time limit.
Users may finish the task before the allocated time expires.

Our previous experiments have anecdotally given us insight on the type of language that participants use in response to correct and incorrect robot behaviors.
For this work, we focus on two types of user speech, (1) correction/clarification, and (2) acceptance, which define negative and positive reward for the preference-learning algorithm respectively.

Corrections and clarifications are reflected by any speech that re-issues or provides additional detail in regards to a previously-issued instruction.
An example of a clarification would be a user saying, ``Robot turn to face the doorway,'' observing the robot perform the action, and then following up the previous command with ``Robot turn 15 degrees to the right,'' which is a minor modification of the initial command signifying the the robot did not perform the desired action at first.
An example of user acceptance of the performed robot action could be the user explicitly acknowledging the correct behavior or moving on to the next subtask.
Verbal commands can add a level of ambiguity when there is subtext to be inferred, so for now we are restricting the language for user clarification/corrections or approval of the robot's behavior to explicitly unambiguous language such as ``No, robot do X'' and ``Yes, robot do Y next''.

\subsubsection{Evaluation}
Evaluation of the user-specific models will be both qualitative and quantitative.
Quantitative metrics will focus on whether the robot was able to accomplish the given objectives from user interactions, the time required to complete the task, and the number of verbal corrections or clarifications a participant gives to the robot in order to perform correct behaviors.
Qualitative metrics will be based on post-interaction surveys where participants answer questions about the robot's movement and overall satisfaction with how the robot performed.

\subsection{Message Bridge}
One key component that we have incrementally developed over the past few years in our human-robot dialogue experiments is a message bridge to handle communication between automatic speech recognition (ASR) software and ROS.
The current pipeline used in these previous experiments works as follows: (1) audio from a person is picked up by a push to talk microphone, (2) this speech is processed by an ASR module to convert it to text, (3) the processed text is then forwarded to a classifier inside a natural language processing (NLP) component to infer the underlying intent of the speech based on a corpus of previously collected data (e.g., convert ``Okay robot move ahead two feet'' to ``Move two feet forward''), (4) the processed text is converted to a ROS compatible format which is then transmitted over the message bridge, and (5) on the ROS side the robot receives the instruction and executes the behavior.

The proposed preference learning component would be inserted into this pipeline as the penultimate stage, where the message received over the bridge is processed by the algorithm which then determines which of the possible navigation trajectories allowed by the instruction would be the one most preferred by the user.
Once this choice is made, then this preference component would then forward the command to the robot to execute.
The message bridge we have developed is also bidirectional with text-to-speech capabilities and would allow the robot to give direct feedback to the user for an enhanced interaction.

\subsection{Waypoints}
The eight navigation instruction types listed in Table \ref{tab:instructions} all result in the robot moving to some waypoint in the environment.
We represent these waypoints using ``poses'' in ROS, with one field being a three-dimensional point in space, and another field being a quaternion to signify orientation.
Our assumption at this point is that these waypoints represent the final position of the robot after executing a command from the user, and that we can observe the preference-learning algorithm adapting to user preferences by evaluating how the position and orientation changes over time for instructions in a similar class.
In other words, we are not currently concerned about \textit{how} the robot navigates to the waypoint as it travels through the environment (other than it not hitting obstacles of course), but only about its \textit{end location} after executing an instruction.
Focusing on just end points for navigation, in combination with the limited space in our virtual environments, should ideally reduce the amount of training required for the robot to exhibit the desired behavior using reinforcement learning, a lingering problem in the research area that partially motivated the work in \cite{michini}.

The details for the algorithm training process are still a work in progress, but we have identified a set of features to use to represent the robot's state.
The first subset of features covers the robot's position and orientation in space as described above.
The second subset of features describes the space around the robot to signify its visibility and immediate operating space; for example, a waypoint that has a robot facing a wall $<$1ft directly in front of it would have less visibility and room for operation compared to a robot in the same 3D position but facing open space and being able to see its surroundings from the onboard RGB camera.

\section{Limitations and Future Work}
The current approach assumes predefined waypoints as landmarks. For instance, ``Go to the doorway'' would require the robot to cross-reference the landmark in its immediate physical context, which is beyond the scope of this work. We are exploring multimodal approaches that use robot localization and perception processing to automatically resolve such references. This access to streaming data would allow for more flexibility by allowing the user to shift between landmark-based on metric-based instructions.

\subsection{Future Extensions}
\label{sec:future}
As mentioned earlier, high level language from the user introduces ambiguity when trying to determine positive and negative feedback, where the algorithm would have to take into account the current navigation subtask to determine if commands from the user are directed towards the action that was just performed or a subsequent one.
A similar problem can be introduced even when the user wants to signify correct robot behavior where the robot actually performs the desired behavior and the participant immediately moves on to the next subtask, such as taking a picture.
For example, if a user gives the robot the command to navigate to a location ``Go to the room in front of you'' and then the following command is something along the lines of ``Turn to face [nearby object]'', it is difficult to tell if the second command is a modification of the first one (signifying incorrect robot behavior), or the beginning of the next subtask.
Essentially, without explicit acknowledgements of correct or incorrect robot behavior from the user, great care must be taken to process user feedback as positive or negative.
One approach to address this problem is the use of sentiment analysis (e.g., Yang et al. \shortcite{NIPS2019_8812}) to determine positive or negative feedback.

\section{Summary}
Effective human teams are ones where team members can automatically recognize or infer the abilities, preferences, and situational awareness of other teammates to predict how a task will be executed.
Providing robot teammates with this capability will help decrease user burden by requiring fewer user interventions to adjust robot behaviors to what is desired.
In this paper, we discuss the initial steps and immediate planned work to implement preference-based reinforcement learning in an exploration task where a user gives verbal commands to a robot and uncovers information about an unknown environment through the robot's sensors.
The planned outcome of this work is an autonomous navigation algorithm that continuously makes adjustments to its navigation behavior based on its growing interaction history with the user.

\section{Acknowledgements}
The authors would like to thank John Rogers III for his help and the Director's Research Awards committee at the Army Research Laboratory for funding this research.

\bibliography{main}

\begin{thebibliography}{}

\bibitem[\protect\citeauthoryear{Bonial \bgroup et al\mbox.\egroup
  }{2017}]{bonialAAAI}
Bonial, C.; Marge, M.; Foots, A.; Gervits, F.; Hayes, C.; Henry, C.; Hill, S.;
  Leuski, A.; Lukin, S.; Moolchandani, P.; Pollard, K.; Traum, D.; and Voss, C.
\newblock 2017.
\newblock Laying down the yellow brick road: Development of a wizard-of-oz
  interface for collecting human-robot dialogue.
\newblock {\em AAAI Fall Symposium Series: Natural Communication for
  Human-Robot Collaboration}.

\bibitem[\protect\citeauthoryear{Desai \bgroup et al\mbox.\egroup
  }{2013}]{desaikania}
Desai, M.; Kaniarasu, P.; Medvedev, M.; Steinfeld, A.; and Yanco, H.
\newblock 2013.
\newblock Impact of robot failures and feedback on real-time trust.
\newblock {\em 8th ACM/IEEE International Conference on Human-Robot
  Interaction}  251--258.

\bibitem[\protect\citeauthoryear{Desai, Medvedev, and Vazquez}{2012}]{Desai12}
Desai, M.; Medvedev, M.; and Vazquez, M.
\newblock 2012.
\newblock Effects of changing reliability on trust of robot systems.
\newblock In {\em ACM/IEEE International Conference on Human-Robot
  Interaction},  73--80.

\bibitem[\protect\citeauthoryear{Hancock \bgroup et al\mbox.\egroup
  }{2011}]{hancock}
Hancock, P.; Billings, D.; Schaefer, K.; Chen, J.; de~Visser, E.; and
  Parasuraman, R.
\newblock 2011.
\newblock A meta-analysis of factors affecting trust in human-robot
  interaction.
\newblock {\em Human Factors: The Journal of Human Factors and Ergonomics
  Society}  517--527.

\bibitem[\protect\citeauthoryear{Knox and Stone}{2012}]{knox12}
Knox, W., and Stone, P.
\newblock 2012.
\newblock Reinforcement learning from simultaneous human mdp reward.
\newblock In {\em 11th International Conference on Autonomous Agents and
  Multiagent Systems}.

\bibitem[\protect\citeauthoryear{Lukin \bgroup et al\mbox.\egroup
  }{2018}]{lukinACL}
Lukin, S.; Gervits, F.; Hayes, C.; Leuski, A.; Moolchandani, P.; Rogers, J.;
  Amaro, C.; Marge, M.; Voss, C.; and Traum, D.
\newblock 2018.
\newblock Scoutbot: A dialogue system for collaborative navigation.
\newblock {\em 56th Annual Meeting of the Association for Computational
  Linguistics - Demonstrations}.

\bibitem[\protect\citeauthoryear{Marge \bgroup et al\mbox.\egroup
  }{2016}]{Marge2016}
Marge, M.; Bonial, C.; Byrne, B.; Cassidy, T.; Evans, A.~W.; Hill, S.~G.; and
  Voss, C.
\newblock 2016.
\newblock {A}pplying the {W}izard-of-{O}z {T}echnique to {M}ultimodal
  {H}uman-{R}obot {D}ialogue.
\newblock In {\em Proc. of RO-MAN}.

\bibitem[\protect\citeauthoryear{Marge \bgroup et al\mbox.\egroup
  }{2017}]{margeNLP}
Marge, M.; Bonial, C.; Foots, A.; Hayes, C.; Henry, C.; Pollard, K.; Artstein,
  R.; Voss, C.; and Traum, D.
\newblock 2017.
\newblock Exploring variation of natural human commands to a robot in a
  collaborative navigation task.
\newblock {\em Proceedings of the First Workshop on Language Grounding for
  Robotics}  58--66.

\bibitem[\protect\citeauthoryear{Michini \bgroup et al\mbox.\egroup
  }{2015}]{michini}
Michini, B.; Walsh, T.; Agha-Mohammadi, A.; and How, J.
\newblock 2015.
\newblock Bayesian nonparametric reward learning from demonstration.
\newblock {\em IEEE Transactions of Robotics} 31(2):369--386.

\bibitem[\protect\citeauthoryear{Moolchandani, Hayes, and
  Marge}{2018}]{moolchandani}
Moolchandani, P.; Hayes, C.; and Marge, M.
\newblock 2018.
\newblock Evaluating robot behavior in response to natural language.
\newblock {\em Companion of the 2018 ACM/IEEE International Conference on
  Human-Robot Interaction}  197--198.

\bibitem[\protect\citeauthoryear{Riek}{2012}]{riek2012}
Riek, L.
\newblock 2012.
\newblock Wizard of oz studies in hri: A systematic review and new reporting
  guidelines.
\newblock {\em Journal of Human-Robot Interaction} 1(1).

\bibitem[\protect\citeauthoryear{Wirth \bgroup et al\mbox.\egroup
  }{2017}]{Wirth17}
Wirth, C.; Akrour, R.; Neumann, G.; and Furnkranz, J.
\newblock 2017.
\newblock A survey of preference-based reinforcement learning methods.
\newblock In {\em Journal of Machine Learning Research}, volume~18,  1--46.

\bibitem[\protect\citeauthoryear{Yang \bgroup et al\mbox.\egroup
  }{2019}]{NIPS2019_8812}
Yang, Z.; Dai, Z.; Yang, Y.; Carbonell, J.; Salakhutdinov, R.~R.; and Le, Q.~V.
\newblock 2019.
\newblock Xlnet: Generalized autoregressive pretraining for language
  understanding.
\newblock In Wallach, H.; Larochelle, H.; Beygelzimer, A.; d~Alch\'{e}-Buc, F.;
  Fox, E.; and Garnett, R., eds., {\em Advances in Neural Information
  Processing Systems 32}. Curran Associates, Inc.
\newblock  5753--5763.

\bibitem[\protect\citeauthoryear{Ziebart \bgroup et al\mbox.\egroup
  }{2008}]{ziebart}
Ziebart, B.; Maas, A.; Bagnell, J.; and Dey, A.
\newblock 2008.
\newblock Maximum entropy inverse reinforcement learning.
\newblock {\em Proceedings of the 23rd AAAI Conference on Artificial
  Intelligence} 3:1433--1438.

\end{thebibliography}
\bibliographystyle{aaai}

\end{document}